\documentclass{article}

\usepackage{arxiv}

\usepackage[utf8]{inputenc} 
\usepackage[T1]{fontenc}    
\usepackage{hyperref}       
\usepackage{url}            
\usepackage{booktabs}       
\usepackage{amsfonts}       
\usepackage{nicefrac}       
\usepackage{microtype}      
\usepackage{lipsum}
\usepackage{fancyhdr}       
\usepackage{graphicx}       

\usepackage{stfloats}
\usepackage{verbatim}
\usepackage{subcaption}
\graphicspath{ {./images} }
\usepackage{caption}
\usepackage{amsmath}
\usepackage{tabularx}
\usepackage[normalem]{ulem}
\usepackage{algorithm}
\title{A Simple Attention-Based Mechanism for Bimodal Emotion Classification}

\author{
  Mazen Elabd, Sardar Jaf \\
  University of Sunderland \\
  Sunderland \\
  \texttt{\{mazen.elabd, sardar.jaf\}@sunderland.ac.uk} \\
}

\begin{document}

\maketitle

\begin{abstract}
Big data contain rich information for machine learning algorithms to utilize when learning important features during classification tasks. Human beings express their emotion using certain words, speech (tone, pitch, speed) or facial expression. Artificial Intelligence approach to emotion classification are largely based on learning from textual information. However, public datasets containing text and speech data provide sufficient resources to train machine learning algorithms for the tack of emotion classification. In this paper, we present novel bimodal deep learning-based architectures enhanced with attention mechanism trained and tested on text and speech data for emotion classification. We report details of different deep learning based architectures and show the performance of each architecture including rigorous error analyses. Our finding suggests that deep learning based architectures trained on different types of data (text and speech) outperform architectures trained only on text or speech. Our proposed attention-based bimodal architecture outperforms several state-of-the-art systems in emotion classification.
\end{abstract}

\section{Introduction}
\label{sec.introduction}
Recognizing human emotions automatically from text is a challenging task. Textual data lacks emotional cues, such as speech tone, pitch, vocal expression, facial expression, etc., that are helpful in determining the emotion of a person accurately. Therefore, approaches to automated emotion recognition that rely only on text are inherently limited.  Recent attempts at emotion recognition are directed at using other types of information such as audio, image, and/or video as well as text in order to enrich the information needed to accurately recognize/classify human emotions.

With advances in machine learning applications for various tasks (such as image processing, computer vision, and natural language processing), it is possible to design multimodal emotion recognition/classification systems by training machine learning algorithms on labeled datasets containing different types of data modalities such as text, images, audio, and/or video. Multimodal emotion recognition/classification involves training one or more machine learning algorithms that are best suited for learning from different types of data. For example, we could train algorithm $x$ on text data, $y$ on audio data, and $z$ on visual data and then combine their learning capabilities from these different types of data to perform the classification task. In this paper, we propose a novel design of deep learning architecture that uses text and audio information for the task of emotion classification. We make the following contributions:
\begin{itemize}
    \item  We propose a novel deep-learning multimodal architecture to extract and utilize important features from different types of data (text and audio) to classify emotions into one of several emotion classes.
    \item We propose a new state-of-the-art multimodal emotion classifier.
    \item We propose mid-level data fusion methods for extracting rich features from different unimodal architectures.
\end{itemize}

The rest of the paper is organized as follows: Section~\ref{sec:literature} outlines key related work to our proposed system. We describe our methodology in section~\ref{sec:methodology}, in section~\ref{sec:resultAndduscussion} we show the performance of the proposed system and  compare it against several published works. In section~\ref{sec:Error.Analyses} we analyse the system's performance through several confusion matrices.  We conclude the paper in section~\ref{sec:conclusion}.

\section{Related work}
\label{sec:literature}

The advances in deep learning methods for audio and text processing have motivated researchers to develop different approaches to emotion classification. Generally, these approaches involve training deep learning algorithms on audio and text data, and incorporating a fusion mechanism to combine both modalities. Early feature extraction methods from text, such as word2vec, involve learning information from text in relation to word features. Further advancements in feature extraction from text include  bidirectional long short-term memory networks (Bi-LSTM), gated recurrent unit (GRU), and transformers. Furthermore, rich language models containing multi-lingual and contextual information around words were developed such as Bidirectional Encoder Representations from Transformers (BERT)~\cite{bert}, Robustly optimized BERT (RoBERTa)~\cite{roberat}, and GPT~\cite{Radford2019}, etc.

Methods to combine several data modalities (e.g., different types of data such as text, audio, and images) have been effective in emotion classification~\cite{Zhengetal2022}. Multimodal transformers that allow the concatenation of the feature representation from different types of data have replaced previous methods~\cite{tsai2019multimodal}~\cite{SoumyaandSriram2022}, with further advances through multi-view sequential learning models~\cite{Zadehetal2018} and dynamic fusion graph~\cite{bagher-zadeh-etal-2018}   

Dutta et al~\cite{dutta2023hcam} proposed a hierarchical cross-attention model approach using recurrent and co-attention neural networks models by training them on text and audio data. Their first stage involved training utterance-level embedding extractor from the input data (text and audio), which trains their model to classify individual utterances without accounting for the inter-utterance conversational context. The second stage of training involves feeding the utterance-level embedding from the first stage to a bidirectional gated recurrent unit (Bi-GRU) to introduce inter-utterance context to the model. This stage enhances the model with conversational context information. The last stage consists of the fusion of the embedding from different modalities using self-attention and cross-attention mechanisms. These attention mechanisms allow for capturing relationships and dependencies within input sequences (sentences or speech utterances).

~\cite{li-etal-2020-hitrans} proposed a hierarchical transformer-based model (HiTrans) consisting of transformer-based content and a speaker-sensitive model for emotion classification. Their method uses two hierarchical transformers: a BERT model is used as the low-level transformer for generating local utterance representation, and a high-level transformer that takes the output of the low-level transformer as input to make the model sensitive to the global context of the conversation. They integrate a ``pairwise utterance speaker verification'' (PUSV) method to detect whether two utterances belong to the same speaker.

~\cite{Ghosaletal} proposed a Dialogue Graph Convolutional Network (DialogueGCN) model utilizing a graph neural network based approach to emotion classification by leveraging self and inter-speaker dependency of the interlocutors to model conversational context for emotion classification. One of the main benefits of such a method is addressing context propagation issues present in the current RNN-based methods.

~\cite{Mittal2019M3ERMM} proposed a learning-based system for emotion classification using multiple input modalities that combined information from text, facial cues, and speech. Their system seems to pay more attention to reliable cues while suppressing less helpful cues on a per-sample basis by using Canonical Correlation Analysis, which differentiates between effective and ineffective modalities. The major strength of their proposed system is its robustness to sensor noise in any of the individual modalities.

\section{Proposed Method}
\label{sec:methodology}
We propose a supervised machine learning approach for the task of emotion classification. We use a public dataset containing different types of data (text, speech, and video) for the training and evaluation of a multimodal emotion classifier. The benefit of using different types of data is to extract different levels of details and cues for identifying emotions. Different data types contain different features/information for training machine learning algorithms. Due to limited access to computational resources, we only use text and speech data types in this study. We use different deep learning based feature extraction methods (embedding) for learning patterns from text and speech data. For text, we use BERT embedding~\cite{bert} and for speech, we use Audio Spectogram Transformer (AST) embedding~\cite{Gongetal2021}.

\begin{figure}[t!]
    \centering
    \includegraphics[width=0.9\linewidth]{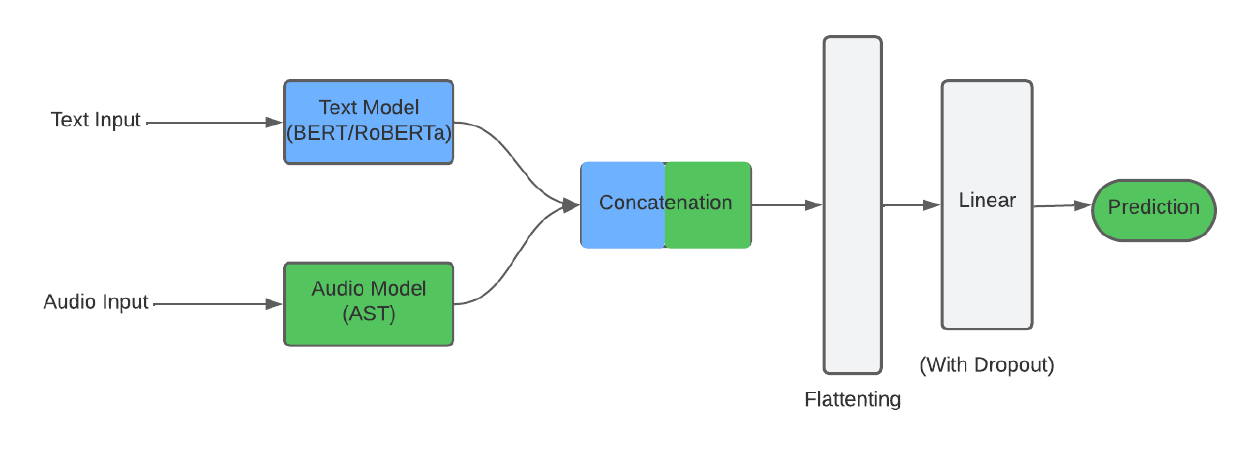}
    \caption{Baseline multimodal classification architecture.}
    \label{fig:baseline-architecture}
\end{figure}

\subsection{System Design}
\label{subsec:system.design}
Our system design is based on fine-tuning attention-based models as well as the reliance on attention-based feature fusion. We believe that two main problems should be addressed when building any multimodal architecture: i) the data representation for each modality in the system (i.e., data representation), and ii) the data concatenation from different modalities (i.e., data fusion).

Our proposed multimodal system for emotion classification relies on the fusion of extracted features from the data through two unimodal architectures: BERT-based architecture for text-based emotion classification, and AST-based architecture for speech-based emotion classification. The last hidden layer from both BERT and AST models is known to have the most contextualised representation of the input sequence. Thus, we specifically use the output of the last hidden layer of fine-tuned versions of BERT and AST to represent the text and audio modalities, respectively. In our attention-based architecture, we adjust the size of the output of the last hidden layer from the text model through a padding layer to match the size of the output of the last hidden layer for the audio model. Similar to late-fusion-based multimodal systems, relying on fine-tuning pre-trained transformer-based models gives the capability of utilising powerful and highly efficient models to represent the multimodal data with a relatively small dataset size and limited hardware resources in the training process.

To address the  feature fusion problem, we rely on a multi-head cross-attention layer as a concatenation technique. If we relied on a method such as averaging or voting, the multimodal system would not benefit from the interactions that could occur between different modalities.

\begin{figure}[t!]
    \centering
    
    \includegraphics[width=0.9\linewidth]{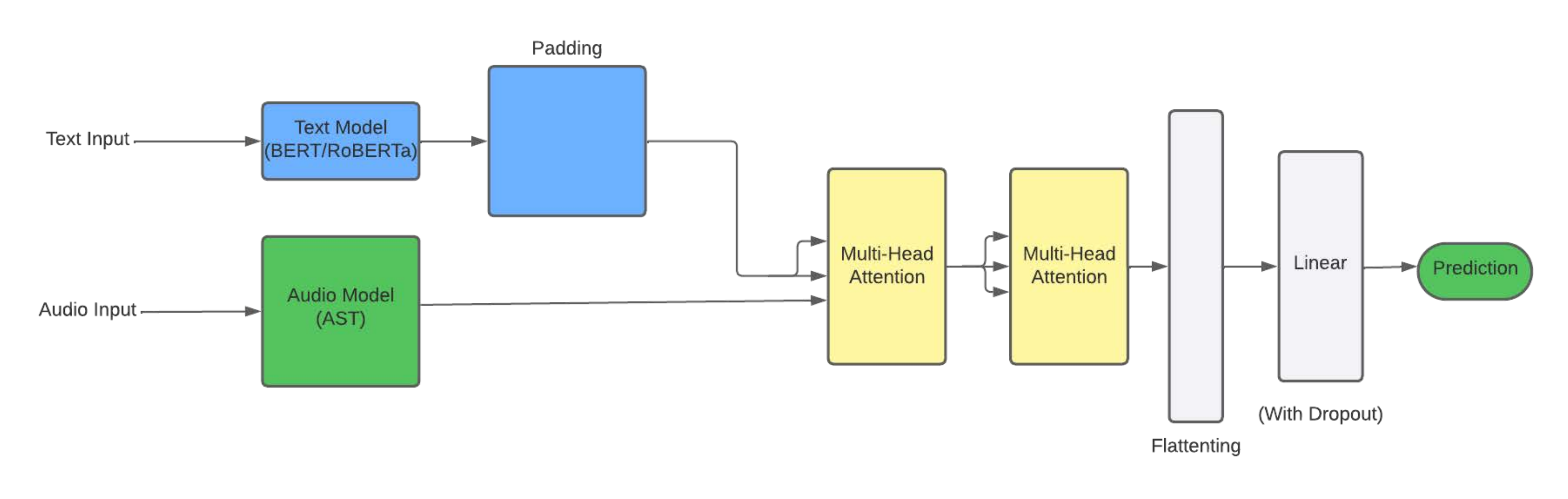}
    \caption{Multi-head attention-based multimodal architecture.}
    \label{fig:cross-attention-architecture}
\end{figure}

Instead of merging both modalities at a classifier level, as done in late feature fusion techniques, we merge them at a data representation layer, which is similar to early fusion techniques, to be able to capture the interactions between audio and text. Our proposed system leverages both early and late fusion techniques. We use powerful pre-trained unimodal models while merging the data before the classifier layer using a simple concatenation layer (as shown in Fig.~\ref{fig:baseline-architecture} for the baseline system) and cross-attention fusion technique, as presented in Fig.~\ref{fig:cross-attention-architecture}. We believe that cross-attention fusion is capable of capturing the interactions between different modalities. It may also be able to make sense of confusing and contradicting data.

To perform the cross-attention fusion, we feed the output from the last hidden layer of the audio model as the key tensor to the multi-head attention layer while feeding the padded output from the last hidden layer of the text model as the query and the value tensors of the multi-head attention layer, as shown in Fig.~\ref{fig:cross-attention-architecture}. The multi-head cross-attention layer is followed by a multi-head self-attention layer which we added in the pursuit of enriching the combined audio and text representation. The combined feature vector is fed as the query, key, and value for the multi-head self-attention layer, as in Fig.~\ref{fig:cross-attention-architecture}. After combining the features from the different data modalities, the combined feature vector undergoes flattening along the first dimension. Then, the combined feature vector is passed to a linear layer with a 10\% dropout, which is followed by the classification layer, as in Fig.~\ref{fig:baseline-architecture} and~\ref{fig:cross-attention-architecture}.

\subsubsection{Selecting the unimodal models}
Since the presented architectures in this work rely on unimodal systems to create feature representations for each modality, the selection of the unimodal system is critical to optimise our system’s performance. We believe that the various unimodal models utilized within a multimodal classification architecture should rely on similar feature representation algorithms to reach a homogeneous feature vector when combining them in the system. 

We choose attention-based unimodal approaches to build the feature vector for each modality since they are proven to achieve state-of-the-art results in many benchmarks and most importantly to maintain homogeneous feature representations between modalities. Attention-based models are popular in various unimodal classification domains, such as text, speech, video, and image, which enable our multimodal systems to easily scale to include another modality. The selected unimodal systems are mainly based on transformer-encoder and only rely on attention to keep the models as homogeneous as possible. In addition, the selected systems are simple and fast to fine-tune.

The last hidden layer from each of the transformers (BERT and AST) contains the most comprehensive and contextualized representation of the input sequence; therefore, it is logical to use it as the feature vector for each modality.

\subsubsection{Baseline system}
 For our baseline model, as depicted in Fig.~\ref{fig:baseline-architecture}, the information derived from the last hidden layer of BERT and AST is fed into a simple concatenation layer as the fusion layer. This was followed by a flattening layer, then a linear layer with dropout (10\% dropout rate) as the classification layer.

\subsubsection{Attention-based system}
For this system, we rely on feeding the feature vector from BERT and AST models to a multi-head cross-attention layer, as in Fig.~\ref{fig:cross-attention-architecture}, which acts as a fusion layer that is capable of capturing the interactions between modalities and developing a comprehensive understanding of the multimodal data. The multi-head cross-attention layer is followed by a multi-head self-attention layer to enrich the combined feature vector before classification. Next, a flattening layer and a linear layer with a 10\% dropout rate serve as the classification layer.

\subsection{dataset}
\label{subsec:dataset}
We use Multimodal EmotionLines Dataset (MELD)~\cite{poriaetal2018.MELD} for the evaluation of the proposed systems. The dataset contains more than over 13000 utterances derived from multi-party conversations. it is divided into three parts: training, validation, and testing. The training set consists of 9,988 utterances, the validation set 1,108 utterances, and the test set 2,610 utterances. We use these partitions unchanged for training, validation, and testing.

\subsection{Hyperparameters}
\label{subsec:experiments-hyperparameters}
The MELD dataset suffers from imbalance class distribution which impacted our decision to rely on AdamW as an optimizer, the weighted cross-entropy as the loss function since the dataset suffer from imbalance class distribution problem, and the weighted F1 score as our main evaluation metric. We integrated early stopping in all of our experiments to stop the training once the weighted F1 score started to decrease. 

Some hyperparameters were only shared across the multimodal experiments such as using a 10\% dropout rate in the last linear layer. Also, both multi-head attention layers used in this work were configured to consist of 128 attention heads\footnote{We carried out experiments on fewer than 128 attention heads, 64 heads and 32 heads, but the obtained results were slightly worse than using 128 attention heads. It could be an indication that a higher number of attention heads than 128 introduces a model that is more capable of capturing the interactions between different modalities.}. We performed limited experiments to reduce the number of attention heads, although the weighted F1 score was negatively impacted.

Despite the fact that we managed to achieve state-of-the-art performance, the selected hyperparameters still might not be perfect since we were only aiming to get an acceptable learning curve. Also, some hyperparameters, such as learning rate and batch size in table Table~\ref{tab:hyperparameters}, were configured due to hardware limitations as we only depended on a 12GB NVIDIA GeForce GPU to carry out the experiments in this work.

\begin{table}[!t]
\caption{The learning rate and the batch size used in each experiment\label{tab:hyperparameters}}
\centering
\begin{tabular}{|l|l|l|}
\hline
Model            & Learning Rate & Batch Size \\ \hline
BERT             & 5e-6          & 2 \\ \hline
AST              & 5e-6          & 2 \\ \hline
Baseline         & 5e-8          & 2 \\ \hline
Attention-based  & 5e-8          & 2 \\ \hline
\end{tabular}
\end{table}

\subsubsection{evaluation metric}
Some of the most robust and widely used performance measures for classification tasks are: (i) recall, (ii) precision, and (iii) F1 score. The recall metric measures the proportion of actual positives that the model is able to classify. Precision, on the other hand, measures the proportion of predicted positives that are correctly identified. Individually, these metrics do not offer a complete view of model performance. The F1 score computes the harmonic mean of the recall and precision offering a robust evaluation. Since the data are imbalanced, we use a weighted F1 score to account for each class's contribution based on its support, which considers the number of actual occurrences of the class in the dataset. Moreover, we compute and present the macro average and weighted average performances of the models to show the overall system performance. We present the macro average to assess precision, recall, and F1 score across individual classes, treating all classes equally. We also use the weighted average performance to account for the importance/weight of each category when calculating the overall system performance.

 \begin{table}[t!]
    \caption{Performance of text and audio unimodal models by emotion class.\label{tab:unimodal.evaluation.per.category.MELD.emotion}}
    \centering
      \resizebox{\columnwidth}{!}{
    \begin{tabular}{|l|l|l|l|l|l|l|}
    \hline
     & \multicolumn{3}{c}{text} & \multicolumn{3}{c}{speech} \\ \hline
     
        Categories & precision & recall & F1 score  & precision & recall & F1 score \\ \hline
        Anger & 0.497 & 0.487 & 0.492 & 0.387 & 0.249 & 0.303 \\ \hline
        Disgust & 0 & 0 & 0  & 0 & 0 & 0 \\ \hline
        Fear & 0.5 & 0.02 & 0.038 & 0 & 0 & 0\\ \hline
        Joy & 0.602 & 0.627 & 0.615 & 0.228 & 0.192 & 0.208 \\ \hline
        Neutral & 0.741 & 0.847 & 0.790 & 0.577 & 0.757 & 0.655 \\ \hline
        Sadness & 0.376 & 0.337 & 0.355 & 0.175 & 0.154 & 0.164 \\ \hline
        Surprise & 0.654 & 0.537 & 0.590  & 0.314 & 0.231 & 0.266 \\ \hline
        \textbf{Macro Avg.} & \textbf{0.481} & \textbf{0.408} & \textbf{0.412} & \textbf{0.240} & \textbf{0.226} & \textbf{0.228} \\ \hline
        \textbf{Weighted Avg.} & \textbf{0.623} & \textbf{0.654} & \textbf{0.633} & \textbf{0.411} & \textbf{0.464} & \textbf{0.429} \\ \hline
    \end{tabular}
    }
\end{table}

\section{Result and Discussion}
\label{sec:resultAndduscussion}
The proposed deep learning systems consist of appropriate deep learning algorithms for each data type. They are trained and tested on two types of data: text and speech data. 

We present  several performance aspects of the proposed systems for recognizing different types of emotions. We use several standard evaluation metrics for each category: precision, recall and F1 score, macro average and weighted average F1 score. These performance measures allow us to identify the model's performance at the category level. In the following subsections, we outline the performance of the systems when trained and tested for emotion classification.

\subsection{Model evaluation result at category level}

\subsubsection{Text unimodal performance}
Table~\ref{tab:unimodal.evaluation.per.category.MELD.emotion} shows the performance of the text-based unimodal system in classifying emotions when trained and tested on text data. The second, third, and fourth columns show the precision, recall and F1 score for each type of emotion. The text-based unimodal model failed to classify the ``disgust'' emotion. Although it scores a recall of 50\% when classifying the ``fear'' category, it has a very poor precision of 0.02 (2\%), resulting in a very poor F1 score of 0.038. The text-based unimodal model classifies the ``neutral'' category more accurately than other categories, achieving an F1 score of 0.790. This is followed by the ``joy'' category (with an F1 score of 0.615) and the ``surprise'' category (with an F1 score of 0.590). The performance in classifying ``anger'' and ``sadness'' categories are F1 scores of 0.492 and 0.355, respectively.

\subsubsection{Speech unimodal performance} 
The speech-based unimodal system performance is relatively similar to that of the text-based unimodal system. It performs better in classifying the ``neutral'' emotion than any other type of emotion and performs very poorly in classifying ``disgust'' and   ``fear'' emotions. As presented in Table~\ref{tab:unimodal.evaluation.per.category.MELD.emotion}, the speech-based unimodal system fails to classify ``disgust'' and ``fear'' categories while achieving an F1 score of over 0.655 for the ``neutral'' category. Similar to the text-based unimodal system, following the ``neutral'' category, the speech unimodal system achieves an F1 score of 0.303, 0.266, and 0.208 for the ``anger'', ``surprise'', and ``joy'' categories, respectively.

\begin{table}[t!]
    \caption{Performance of the baseline multimodal model by emotion class. \label{tab:baseline.bimodal.evaluation.per.category.MELD.emotion}}
    \centering
    \small
    \begin{tabular}{|l|l|l|l|}
    \hline
        Categories & precision & recall & F1 score \\ \hline
        Anger & 0.577 & 0.435 & 0.496   \\ \hline
        Disgust & 0.5 & 0.012 & 0.029  \\ \hline
        Fear & 0.154 & 0.08 & 0.105  \\ \hline
        Joy & 0.568 & 0.657 & 0.609   \\ \hline
        Neutral & 0.793 &  0.766 & 0.779  \\ \hline
        Sadness & 0.333 & 0.375 & 0.353  \\ \hline
        Surprise & 0.483 & 0.705 & 0.573  \\ \hline
        \textbf{Macro Avg.} & \textbf{0.487} & \textbf{0.433} & \textbf{0.421}  \\ \hline
        \textbf{Weighted Avg.} & \textbf{0.640} & \textbf{0.635} & \textbf{0.627}   \\ \hline
    \end{tabular}
\end{table}

\begin{table}[t]
    \caption{Performance of the attention-based multimodal model by emotion class.\label{tab:system.evaluation.per.category.MELD}}
    \centering
    \begin{tabular}{|l|l|l|l|}
    \hline
        Categories & precision & recall & F1 score \\ \hline
        Anger & 0.516 & 0.672 & 0.584 \\ \hline
        Disgust & 0 & 0 & 0 \\ \hline
        Fear & 0 & 0 & 0 \\ \hline
        Joy & 0.675 & 0.709 & 0.692 \\ \hline
        Neutral & 0.858 & 0.792 & 0.824 \\ \hline
        Sadness & 0.478 & 0.418 & 0.446 \\ \hline
        Surprise & 0.616 & 0.868 & 0.721 \\ \hline
        \textbf{Macro Avg.} & \textbf{0.449} & \textbf{0.494} & \textbf{0.467} \\ \hline
        \textbf{Weighted Avg.} & \textbf{0.689} & \textbf{0.706} & \textbf{0.693} \\ \hline
    \end{tabular}
\end{table}

\subsubsection{Baseline multimodal performance}
We evaluated the proposed multimodal baseline system to assess its performance at classifying different types of emotions. 
Table~\ref{tab:baseline.bimodal.evaluation.per.category.MELD.emotion} presents the performance of the baseline system. The system produced the highest F1 score in classifying the ``neutral'' emotion (0.779), followed by ``joy'' (0.609).  The system produced the lowest F1 scores in classifying the ' disgust'' and `fear'' categories with 0.029 and 0.105, respectively. The overall F1 score across all the emotion categories is 0.627. The baseline system has a reasonable overall weighted average of precision and recall across all emotion categories, producing scores of 0.640 and 0.635, respectively.

\subsubsection{Attention-based multimodal performance}
Table \ref{tab:system.evaluation.per.category.MELD} presents the performance of the attention-based system. The system produced the highest weighted F1 score, recall, and precision compared to other models in this work, with scores of 0.693, 0.706, and 0.689, respectively. It also produced the highest macro average F1 score, with 0.467. Although it was unable to classify any instances from the least represented classes ``fear'' and ``disgust'', the system produced the highest F1 score in classifying the ``neutral'' emotion (0.824), followed by ``surprise'' (0.721), and then ``joy'' (0.692). 

\begin{table}[t!]
\caption{Comparative result.  Weighted F1 score (in \%)\label{tab:system.evaluation.on.MELD.emotion}}
    \centering
    \begin{tabular}{|l|l|l|l|}
    \hline
        \textbf{Models} & \textbf{Speech} & \textbf{Text} & \textbf{Speech+Text}  \\ \hline
        Majumder et al~\cite{Majumderetal2018} & 41.80 & 57.00 & 60.30  \\ \hline
        Ho et al~\cite{Hoetal2020} & 45.30 & 59.00 & 60.50 \\ \hline
        Lian et al~\cite{ZLianetal2021} & 38.20 & 58.30 & 60.50 \\ \hline
        Baijun et al~\cite{Baijunetal2021} & 32.10 & 61.20 & 64.00 \\ \hline
        Dutta et al~\cite{dutta2023hcam} & 50.10 & 65.60 & 65.80  \\ \hline
        \textbf{Proposed baseline model} & 42.90 &  63.3  & 62.7  \\ \hline
        \textbf{Proposed attention-based model}& 42.90 &  63.3  & \textbf{69.7}  \\ \hline
    \end{tabular}
\end{table}

\subsection{System performance comparison with previous works}
Table~\ref{tab:system.evaluation.on.MELD.emotion} presents the weighted F1 score performance of our proposed systems (baseline and attention-based) and several previously published works on unimodal and multimodal emotion classification. Our proposed unimodal systems lagged behind some previously published systems, while our proposed multimodal systems have performed better than most of the other systems.

Our proposed attention-based system significantly outperformed state-of-the-art systems by achieving a 69.7\% F1 score, nearly 5\% higher than the reported state-of-the-art system proposed by \cite{dutta2023hcam}. This is despite the low performance of our unimodal systems compared to the unimodal systems proposed by~\cite{dutta2023hcam}. This indicates the strength of our novel method of combining the last hidden layers of attention-based text and speech unimodal models using a multi-head cross-attention layer followed by a multi-head self-attention layer for fusing text and audio modalities.

\subsection{Multimodal performance discussion}
Our proposed speech-based unimodal emotion classifier underperforms compared to those classifiers proposed by Dutta et al~\cite{dutta2023hcam} and Ho et al~\cite{Hoetal2020}. Our text-based unimodal model is only outperformed by the model proposed by Dutta et al~\cite{dutta2023hcam}. However, our multimodal classifier outperforms the state-of-the-art. We anticipate that the superior performance of our multimodal system results from our homogenous attention-based model which includes an attention-based fusion method that merges the learned information from attention-based unimodal models, which are fine-tuned versions of BERT and AST. In comparison to several other published works, our speech-based unimodal performance is on par with many other systems and better than others in some cases. Our text-based unimodal model performs better than almost all other published works, as shown in Table~\ref{tab:system.evaluation.on.MELD.emotion}. Furthermore, our multimodal attention-based system outperforms all other published works, making it the state-of-the-art multimodal emotion classifier.

\begin{figure}[t!]
    \centering
         \includegraphics[width=0.97\textwidth]{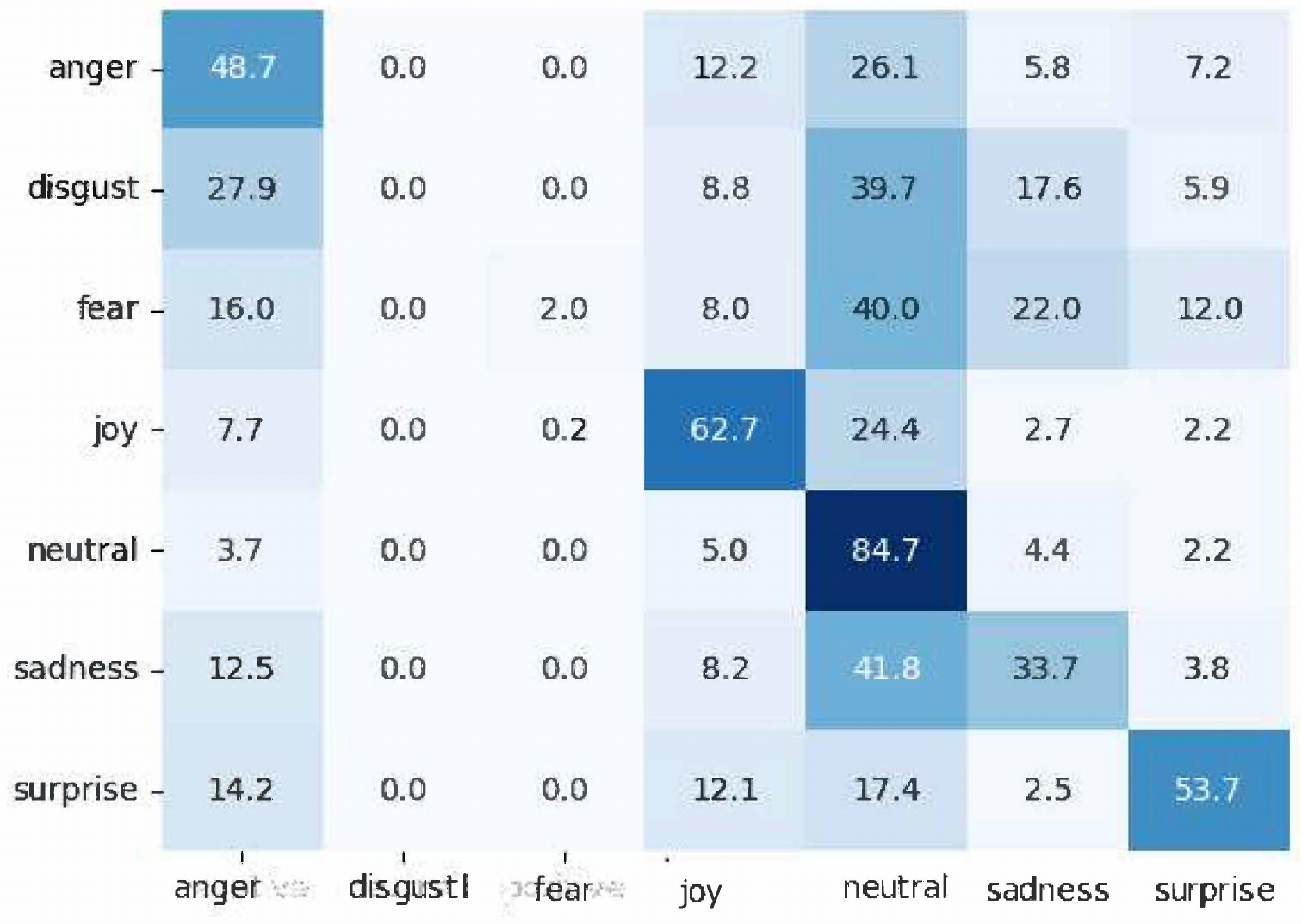}
         \caption{text unimodal model}
         \label{fig:textUniModel}   
\end{figure}

\section{Error analyses: confusion matrix}
\label{sec:Error.Analyses}
Machine learning applications for supervised classification tasks require a set of labeled data that represents the classes. Labeled datasets often contain errors. One of the main issues is label errors (annotation errors), where some data samples are labeled with incorrect emotions.
Annotation errors impact machine learning algorithms' performance because the algorithms learn from the errors present in the annotated dataset. Confusion matrix graphs are helpful for highlighting classification errors by identifying misclassified classes and the classes they are confused with. We present several confusion matrices to outline the classification errors of our proposed unimodal and multimodal systems in classifying different types of emotions.

\subsection{Emotion misclassification errors}
As presented in Fig.~\ref{fig:textUniModel}, our text unimodal system miscalssifies``anger'' class as ``netural'' 26.1\%. The class``disgust'' is misclassified with ``neutral'' 39.7\%, ``fear'' is misclassified with``neutral'' 40\%, and topped by ``sadness'' with 41.8\%. Contrary to the ``neutral'' class that has the largest data samples in the dataset, the text-based unimodal does not confuse any class  (except ``joy'' confused with ``fear'', though negligibly) with the ``disgust'' and ``fear'' classes, which have the  fewest data samples in the dataset.

\begin{figure}[t!]
\centering
        \includegraphics[width=0.7\textwidth]{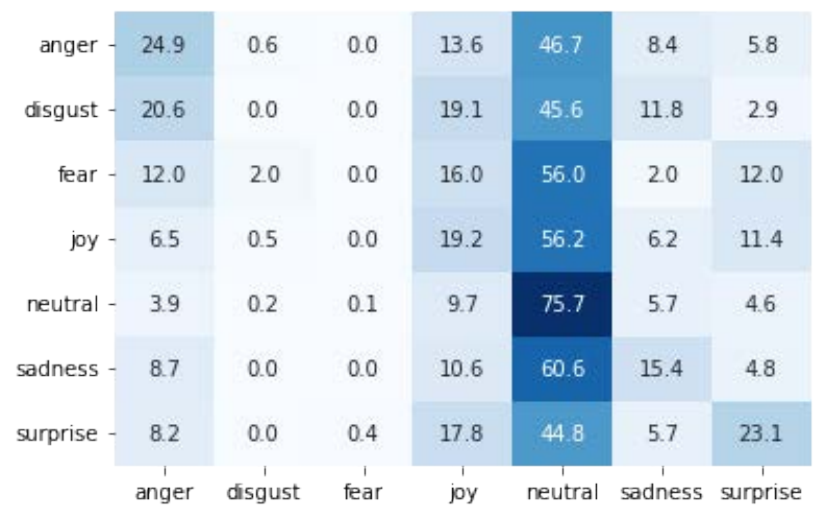}
        \caption{Speech unimodal model}
        \label{fig:speechUniModel}
\end{figure}

\begin{figure}[t!]
\centering
        \includegraphics[width=0.7\textwidth]{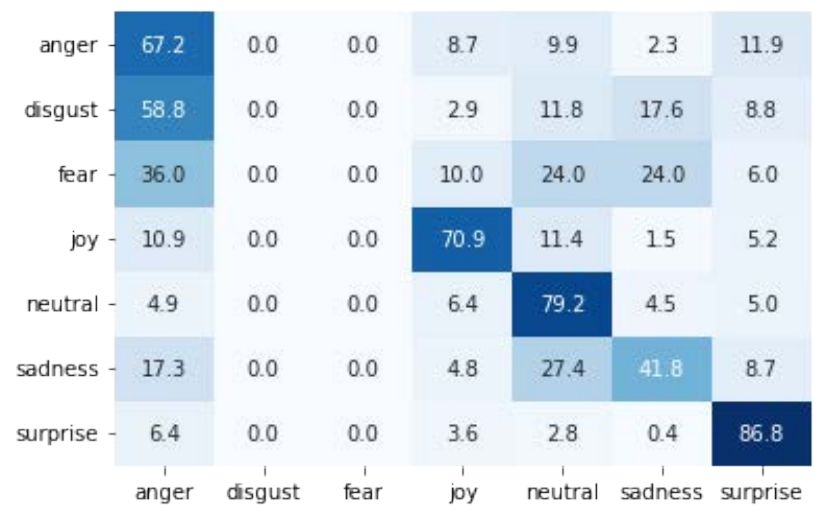}
        \caption{Attention-based multimodal model}
        \label{fig:attentionModel}
\end{figure}

The speech-based unimodal seems to miss-classify most of the classes largely with the ``neutral'', as shown in Fig.~\ref{fig:speechUniModel}. The miss-classification error rate between ``neutral'' and all other class of emotions range from 44.8\%, where ``surprise'' is miss-classified as ``netural'', to 60.0\%, where ``sadness'' is miss-classified as ``neutral''. The ``neutral'' class appears to be the least miss-classified class. It has 75.7\% accuracy. It's highest miss-classified rate of 9.7\% is  with ``joy''  followed by  ``sadness'' (5.7\%) and ``surprised'' (4.6\%).

The cross-attention multimodal system, as presented in Fig.~\ref{fig:attentionModel}, confused ``disgust'' class as ``anger'' (58.8\% error rate), ``neutral'' (11.8\% error rate), ``sadness'' (17.6\% error rate), and ``surprised 8.8\% error rate). The cross-attention multimodal has increased the accuracy of classifying ``anger'' emotion by reducing the error rate of miss-classification of this type of emotion with ``neutral'' to 9.9\% compared with the error rate recorded in the unimodal systems. However, ``anger'' emotion is miss-classified as ``joy'' 8.7\%, as ``sadness'' 2.3\% and as ``surprise'' 11.1\%. 
``Sandess'' emotion class has been confused by the cross-attention multimodal system largely as ``neutral'' with an error rate of 27.4\%, followed by 17.3\% error rate when miss-classified as ``anger'', 4.8\% as ``joy'' and 8.7\% as ``surprise''. Other emotion classes (``joy'' and ``surprise'') are mostly miss-classified as ``neutral'' and``neutral'' class is largely confused as ``joy'' (6.4\% error rate). It appears our proposed multi-head cross-attention system performed best at classifying ``surprise'' emotion with 86.8\% accuracy.

\section{conclusion and future work}
\label{sec:conclusion}
Recent advances in automated data classification of emotion involve the application of machine learning algorithms to automate the process of classifying certain types of emotions (e.g., anger, happy, surprised, etc.). The focus of this study was to design and evaluate novel deep learning architectures  that learn from different types of data (text and speech) to classify different kinds of emotions. We applied deep learning based language models, such as BERT and Audio Spectrogram Transformer. We proposed novel deep learning system to fuse embedding data  using attention mechanism for extracting important features from different types of data to classify emotion. We have measured the performance of each architecture using different performance matrices. We present rigorous error analyzes of the proposed system classification performance, where certain emotion classes are miss-classified. Our finding suggests that deep learning architectures trained only on text or speech data could underperform architectures that are trained on a fusion of data (e.g., text and speech). Thus, proving that multimodal systems could perform better than unimodal systems for emotion classification.  Our future work is to extend our multimodal system to evaluate it on sentiment analyses, and to fuse other data types when training it (e.g., fusing video data with text and speech). We believe this is feasible because the public dataset (MELD) contains videos and our current architecture is designed to be simple and thus allowing us to easily extend it to include information learn by current embedding algorithms from videos. Also, MELD data contains information for sentiment analysis.

\bibliographystyle{unsrt} 
\bibliography{main}

\begin{thebibliography}{10}

\bibitem{bert}
Jacob Devlin, Ming{-}Wei Chang, Kenton Lee, and Kristina Toutanova.
\newblock {BERT:} pre-training of deep bidirectional transformers for language understanding.
\newblock {\em CoRR}, abs/1810.04805, 2018.

\bibitem{roberat}
Yinhan Liu, Myle Ott, Naman Goyal, Jingfei Du, Mandar Joshi, Danqi Chen, Omer Levy, Mike Lewis, Luke Zettlemoyer, and Veselin Stoyanov.
\newblock Roberta: {A} robustly optimized {BERT} pretraining approach.
\newblock {\em CoRR}, abs/1907.11692, 2019.

\bibitem{Radford2019}
Alec Radford, Jeff Wu, Rewon Child, David Luan, Dario Amodei, and Ilya Sutskever.
\newblock Language models are unsupervised multitask learners.
\newblock 2019.

\bibitem{Zhengetal2022}
Zheng Lian, Bin Liu, and Jianhua Tao.
\newblock Smin: Semi-supervised multi-modal interaction network for conversational emotion recognition.
\newblock {\em IEEE Transactions on Affective Computing}, pages 1--1, 2022.

\bibitem{tsai2019multimodal}
Yao-Hung~Hubert Tsai, Shaojie Bai, Paul~Pu Liang, J.~Zico Kolter, Louis-Philippe Morency, and Ruslan Salakhutdinov.
\newblock Multimodal transformer for unaligned multimodal language sequences, 2019.

\bibitem{SoumyaandSriram2022}
Soumya Dutta and Sriram Ganapathy.
\newblock Multimodal transformer with learnable frontend and self attention for emotion recognition.
\newblock In {\em ICASSP 2022 - 2022 IEEE International Conference on Acoustics, Speech and Signal Processing (ICASSP)}, pages 6917--6921, 2022.

\bibitem{Zadehetal2018}
Amir Zadeh, Paul~Pu Liang, Navonil Mazumder, Soujanya Poria, Erik Cambria, and Louis{-}Philippe Morency.
\newblock Memory fusion network for multi-view sequential learning.
\newblock {\em CoRR}, abs/1802.00927, 2018.

\bibitem{bagher-zadeh-etal-2018}
AmirAli Bagher~Zadeh, Paul~Pu Liang, Soujanya Poria, Erik Cambria, and Louis-Philippe Morency.
\newblock Multimodal language analysis in the wild: {CMU}-{MOSEI} dataset and interpretable dynamic fusion graph.
\newblock In {\em Proceedings of the 56th Annual Meeting of the Association for Computational Linguistics (Volume 1: Long Papers)}, pages 2236--2246, Melbourne, Australia, July 2018. Association for Computational Linguistics.

\bibitem{dutta2023hcam}
Soumya Dutta and Sriram Ganapathy.
\newblock Hcam -- hierarchical cross attention model for multi-modal emotion recognition, 2023.

\bibitem{li-etal-2020-hitrans}
Jingye Li, Donghong Ji, Fei Li, Meishan Zhang, and Yijiang Liu.
\newblock {H}i{T}rans: A transformer-based context- and speaker-sensitive model for emotion detection in conversations.
\newblock In {\em Proceedings of the 28th International Conference on Computational Linguistics}, pages 4190--4200, Barcelona, Spain (Online), December 2020. International Committee on Computational Linguistics.

\bibitem{Ghosaletal}
Deepanway Ghosal, Navonil Majumder, Soujanya Poria, Niyati Chhaya, and Alexander~F. Gelbukh.
\newblock Dialoguegcn: {A} graph convolutional neural network for emotion recognition in conversation.
\newblock {\em CoRR}, abs/1908.11540, 2019.

\bibitem{Mittal2019M3ERMM}
Trisha Mittal, Uttaran Bhattacharya, Rohan Chandra, Aniket Bera, and Dinesh Manocha.
\newblock M3er: Multiplicative multimodal emotion recognition using facial, textual, and speech cues.
\newblock In {\em AAAI Conference on Artificial Intelligence}, 2019.

\bibitem{Gongetal2021}
Yuan Gong, Yu{-}An Chung, and James~R. Glass.
\newblock {AST:} audio spectrogram transformer.
\newblock {\em CoRR}, abs/2104.01778, 2021.

\bibitem{poriaetal2018.MELD}
Soujanya Poria, Devamanyu Hazarika, Navonil Majumder, Gautam Naik, Erik Cambria, and Rada Mihalcea.
\newblock {MELD:} {A} multimodal multi-party dataset for emotion recognition in conversations.
\newblock {\em CoRR}, abs/1810.02508, 2018.

\bibitem{Majumderetal2018}
Navonil Majumder, Soujanya Poria, Devamanyu Hazarika, Rada Mihalcea, Alexander~F. Gelbukh, and Erik Cambria.
\newblock Dialoguernn: An attentive {RNN} for emotion detection in conversations.
\newblock {\em CoRR}, abs/1811.00405, 2018.

\bibitem{Hoetal2020}
Ngoc-Huynh Ho, Hyung-Jeong Yang, Soo-Hyung Kim, and Gueesang Lee.
\newblock Multimodal approach of speech emotion recognition using multi-level multi-head fusion attention-based recurrent neural network.
\newblock {\em IEEE Access}, 8:61672--61686, 2020.

\bibitem{ZLianetal2021}
Zheng Lian, Bin Liu, and Jianhua Tao.
\newblock Ctnet: Conversational transformer network for emotion recognition.
\newblock {\em IEEE/ACM Transactions on Audio, Speech, and Language Processing}, 29:985--1000, 2021.

\bibitem{Baijunetal2021}
Baijun Xie, Mariia Sidulova, and Chung~Hyuk Park.
\newblock Robust multimodal emotion recognition from conversation with transformer-based crossmodality fusion.
\newblock {\em Sensors}, 21(14), 2021.

\end{thebibliography}

\end{document}